\pdfoutput=1

\documentclass[11pt]{article}

\usepackage[final]{acl}

\usepackage{times}
\usepackage{latexsym}
\usepackage{algorithm}
\usepackage{algpseudocode}  
\usepackage{amssymb}
\usepackage[T1]{fontenc}

\usepackage[utf8]{inputenc}

\usepackage{microtype}
\usepackage{pifont} 
\usepackage{soul} 

\usepackage{tipa}
\usepackage{dblfloatfix}
\usepackage{amsmath}
\usepackage{graphicx}
\usepackage{array,booktabs,ragged2e}
\usepackage{multirow}
\usepackage{array}
    \newcolumntype{P}[1]{>{\centering\arraybackslash}p{#1}}
    \newcolumntype{C}[1]{>{\centering\arraybackslash}c{#1}}
    \newcolumntype{L}[1]{>{\centering\arraybackslash}l{#1}}
\usepackage{booktabs}

\newcolumntype{R}[1]{>{\RaggedLeft\arraybackslash}p{#1}}
\newcolumntype{L}[1]{>{\RaggedRight\arraybackslash}p{#1}}
\newcolumntype{M}[1]{>{\centering\arraybackslash}m{#1}}
\usepackage{xcolor}
\definecolor{lpink}{cmyk}{0, 0.7808, 0.4429, 0.1412}
\definecolor{aqua}{cmyk}{0.91, 0, 0.09, 0.36}
\definecolor{ao}{rgb}{0.0, 0.5, 0.0}
\definecolor{amber}{rgb}{1.0, 0.49, 0.0}
\definecolor{dblue}{rgb}{0.0, 0.0, 0.61}
\definecolor{burgundy}{rgb}{0.5, 0.0, 0.13}

\usepackage{inconsolata}

\usepackage{graphicx}
\graphicspath{ {images/} }

\definecolor{dark_blue}{HTML}{4261B3}

\title{Vec2Summ: Text Summarization via Probabilistic Sentence Embeddings}

\author{
    Mao Li \hspace{.5em} Fred Conrad \hspace{.5em} Johann Gagnon-Bartsch \hspace{.5em} \\
    \texttt{\small\{maolee, fconrad, johanngb\}@umich.edu}
}

\begin{document}
\maketitle

\begin{abstract}

We propose Vec2Summ, a novel method for abstractive summarization that frames the task as semantic compression. Vec2Summ represents a document collection using a single mean vector in the semantic embedding space, capturing the central meaning of the corpus. To reconstruct fluent summaries, we perform embedding inversion—decoding this mean vector into natural language using a generative language model. To improve reconstruction quality and capture some degree of topical variability, we introduce stochasticity by sampling from a Gaussian distribution centered on the mean. This approach is loosely analogous to bagging in ensemble learning, where controlled randomness encourages more robust and varied outputs.

Vec2Summ addresses key limitations of LLM-based summarization methods. It avoids context-length constraints, enables interpretable and controllable generation via semantic parameters, and scales efficiently with corpus size—requiring only \( O(d + d^2) \) parameters. Empirical results show that Vec2Summ produces coherent summaries for topically focused, order-invariant corpora, with performance comparable to direct LLM summarization in terms of thematic coverage and efficiency, albeit with less fine-grained detail. These results underscore Vec2Summ’s potential in settings where scalability, semantic control, and corpus-level abstraction are prioritized.

\end{abstract}
\section{Introduction}

Multi‐document summarization seeks to generate concise, coherent synopses from large collections of texts, yet existing paradigms face significant limitations \cite{ma_multi-document_2022}. Extractive methods select and concatenate salient sentences but often yield disjointed summaries that sacrifice overall semantic coherence or omit infrequent yet important content~\cite{nenkova_survey_2012}. Abstractive models—such as pointer–generator networks—produce more fluent summaries but remain bound by fixed context windows, hindering their ability to scale to extensive input sets~\cite{see_get_2017}. More recently, large language models (LLMs) like GPT-4 and PaLM have demonstrated impressive zero‐ and few‐shot summarization performance on benchmarks~\cite{zhang_benchmarking_2023, pu_summarization_2023}, yet these pipelines are computationally expensive, require extensive prompt engineering, and operate as opaque black boxes. Moreover, when applied to unordered corpora (e.g., social media posts or review streams), naive concatenation of inputs can exceed context‐length limits and introduce positional biases—models over‐attend to the beginning and end of long contexts while neglecting the middle~\cite{liu_lost_2024}—and they often hallucinate on high‐diversity collections lacking clear document boundaries~\cite{belem_single_2025}.

Embedding‐based approaches offer an alternative by representing texts as points in a semantic space. The Word2Vec paradigm demonstrated that vector‐space operations capture analogical relationships (e.g., $\mathrm{King} - \mathrm{Man} + \mathrm{Woman} \approx \mathrm{Queen}$)~\cite{mikolov_efficient_2013}, but subsequent analyses revealed that both word and sentence embeddings suffer from \emph{representation degeneration}: high‐frequency tokens cluster near the origin while low‐frequency tokens scatter peripherally, producing narrow, cone‐shaped distributions that undermine isotropy and downstream task performance~\cite{gao_representation_2019, zhang_revisiting_2020}. Pooling strategies (mean‐pooling or [CLS]‐based) yield abstract sentence embeddings whose anisotropy can bias summary selection and degrade coherence.

In this work, we reconceptualize multi‐document summarization as an \emph{information‐compression} problem in embedding space. We observe that the \textbf{mean vector} of a corpus’s sentence embeddings encodes its central semantic tendency, and that probabilistic sampling—via a multivariate Gaussian whose mean is this vector—can enhance embedding‐inversion, yielding more diverse yet coherent reconstructions. We propose \textbf{Vec2Summ}, as illustrated in Figure~\ref{fig:vec2summ}, a framework that:
\begin{figure}
    \centering
    \includegraphics[width=\linewidth]{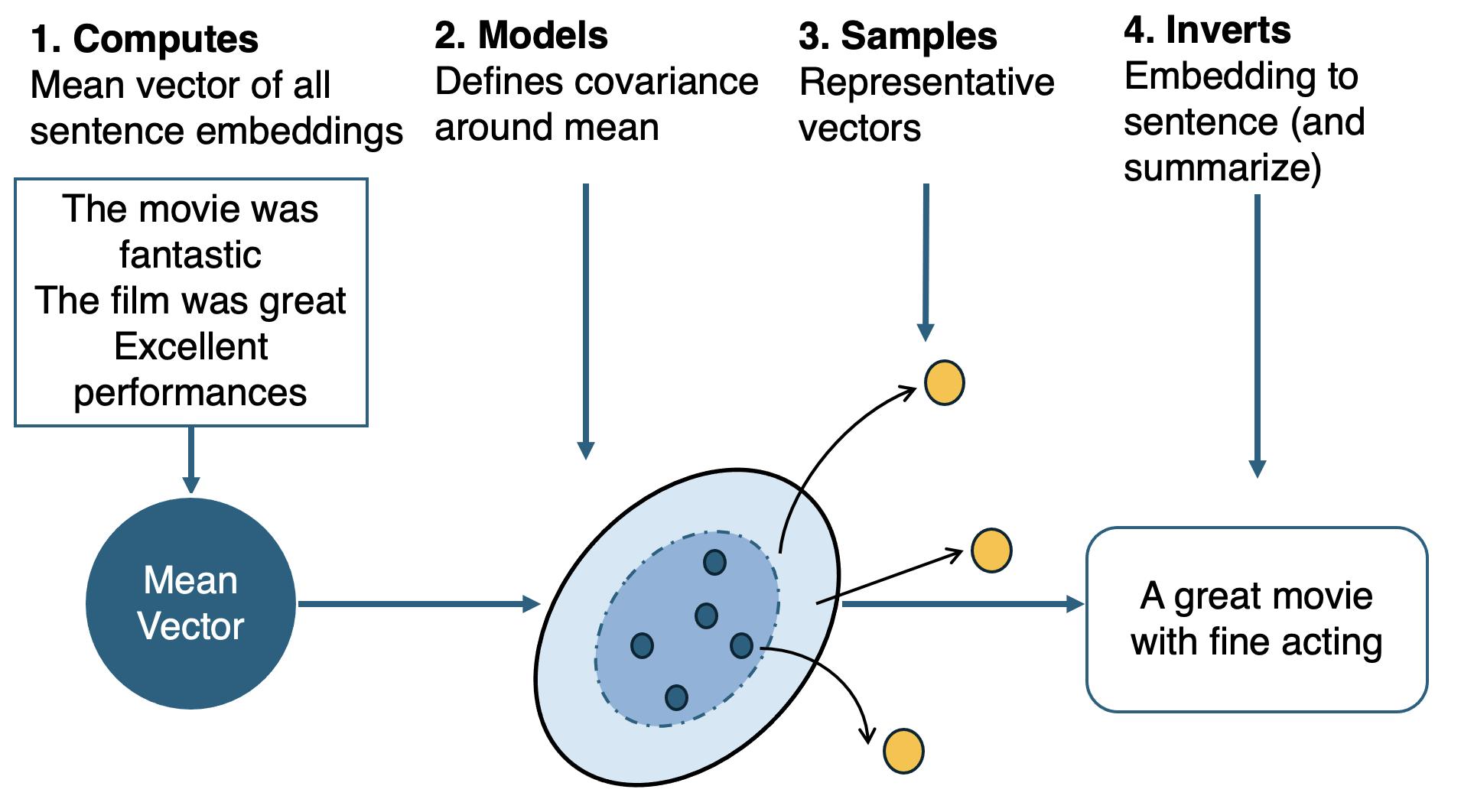}
    \caption{Vec2Summ framework}
    \label{fig:vec2summ}
\end{figure}
\begin{enumerate}
  \item \textbf{Computes} the mean vector of all sentence embeddings to capture the central theme of the corpus.
  \item \textbf{Models} semantic variability by defining a covariance around the mean (used only for sampling to improve inversion diversity).
  \item \textbf{Samples} representative vectors from this Gaussian to introduce controlled perturbations.
  \item \textbf{Inverts} each sampled embedding back to natural language (sentences), producing concise summaries that reflect both core content and thematic variation.
\end{enumerate}

By relying primarily on the mean vector for semantic encoding and using Gaussian sampling as a technique to improve embedding inversion, Vec2Summ overcomes LLM context‐length constraints, captures thematic variability, and generates coherent summaries from arbitrarily large document collections without concatenating all source texts.

\section{Related work}
\subsection{Embedding Spaces and Reconstruction}
Embedding spaces form the theoretical foundation of modern NLP approaches, where words, sentences, and documents are represented as vectors in high-dimensional spaces. The geometric properties of these embedding spaces significantly impact downstream tasks. Mikolov et al.'s pioneering work on Word2Vec \cite{mikolov_efficient_2013} demonstrated that word embeddings capture semantic relationships in vector space, allowing for meaningful vector arithmetic. This discovery fundamentally changed how we represent and manipulate semantic information in natural language processing.

However, research by Gao et al. \cite{gao_representation_2019} and Zhang et al. \cite{zhang_revisiting_2020} revealed that word embeddings are not evenly distributed in high-dimensional spaces—a phenomenon known as representation degeneration. This degeneration is particularly pronounced in pretrained language models like BERT, where embeddings cluster in narrow cones within the vector space, with high-frequency words densely distributed near the origin and low-frequency words sparsely scattered further away. Ethayarajh \cite{ethayarajh_how_2019} demonstrated that in all layers of BERT, ELMo, and GPT-2, word representations are highly anisotropic, occupying a narrow cone in the embedding space rather than being distributed throughout. This anisotropy can significantly impact model performance, as shown by Liang et al. \cite{liang_learning_2021}, who found that BERT embeddings suffer from strong anisotropy that deteriorates their representation capacity.

Studies consistently show that more isotropic (uniformly distributed) embeddings outperform anisotropic ones across various natural language understanding benchmarks \cite{gao_representation_2019}. The challenge becomes even more pronounced when considering sentence-level embeddings, where the semantic interpretation becomes increasingly abstract as word embeddings are pooled. Xu and Koehn \cite{xu_cross-lingual_2021} further revealed that contextual embedding spaces suffer from their natural properties of anisotropy and anisometry, requiring specialized normalization techniques to mitigate these issues.

A critical advancement in embedding research is the ability to reconstruct text from embeddings, which becomes particularly valuable when dealing with anisotropic embedding spaces. Recent work by Morris et al. \cite{morris_text_2023} introduced Vec2Text, a framework for embedding inversion that demonstrates the surprising amount of information preserved in text embeddings despite the representation degeneration problem. Their approach frames the problem as controlled generation: generating text that, when re-embedded, is close to a fixed point in latent space. Vec2Text operates by training two models: a hypothesizer that generates initial text from an embedding, and a corrector that iteratively refines this text to better match the target embedding. This framework has proven effective at recovering original text from its embedding with high fidelity, even in the presence of anisotropic distributions.

While Vec2Text focuses primarily on recovering the original text from its embedding as a reconstruction task, Vec2Summ extends this concept in a novel direction. Rather than simply inverting embeddings to recover source text, Vec2Summ uses embedding inversion to generate diverse summaries that capture the distribution of the original document collection. By modeling the semantic space probabilistically and sampling from this mean and distribution, Vec2Summ leverages embedding inversion as a generative tool for summarization. This approach is particularly powerful because it can work with the inherent anisotropy of pretrained language model embeddings, transforming what is often considered a limitation into an advantage. The probabilistic modeling helps Vec2Summ navigate the narrow cones of the embedding space while still capturing semantic diversity, addressing the limitations of traditional summarization approaches while representing both central tendencies and semantic variability of the source material.

\subsection{Probabilistic Approaches to Summarization}
Probabilistic topic models have a rich history in text summarization, with Latent Dirichlet Allocation (LDA) \cite{blei_latent_nodate} being particularly influential. LDA models documents as mixtures of topics, where each topic is a probability distribution over words. This approach has been extended to summarization tasks by selecting sentences that best represent the dominant topics in a document collection \cite{wang_multi-document_2009}. While these approaches share Vec2Summ's probabilistic foundation, they typically operate in discrete topic spaces rather than continuous embedding spaces.

Gaussian Mixture Models (GMMs) have been applied to various text processing tasks, including document clustering \cite{liu_comparative_2005} and text generation \cite{wang_topic-guided_2019}. Zhu et al. \cite{zhu_texygen_2018} demonstrated that GMMs can effectively model the distribution of sentence embeddings for text generation tasks. Vec2Summ builds on this foundation but focuses specifically on summarization, using a single multivariate Gaussian as a first approximation of the semantic distribution.

Recent advances in diffusion models have revolutionized generative modeling across various domains. Ho et al. \cite{ho_denoising_2020} introduced Denoising Diffusion Probabilistic Models (DDPMs), which have been adapted for text generation by Li et al. \cite{li_diffusion-lm_2022}. These models operate in continuous latent spaces, similar to Vec2Summ's approach. However, while diffusion models typically start from random noise and gradually denoise to generate text, Vec2Summ samples from a learned distribution that captures the semantic structure of the document collection.

\subsection{Sentence Embeddings for Summarization}
Sentence embeddings have become increasingly important for summarization tasks. Reimers and Gurevych \cite{reimers_sentence-bert_2019} introduced Sentence-BERT, which produces semantically meaningful sentence embeddings that can be compared using cosine similarity. More recent work by Ni et al. \cite{ni_large_2021} on General Text Representations (GTR) has further improved sentence embedding quality, providing a solid foundation for Vec2Summ's approach.

A related line of research focuses on selecting representative embeddings for summarization. Kobayashi et al. \cite{kobayashi_summarization_2015} proposed a method for extractive summarization that selects sentences to maximize the coverage of the original document embeddings. Similarly, Chu and Liu \cite{chu_meansum_2019} introduced MeanSum, which generates abstractive summaries by decoding from the mean of review embeddings.

Vec2Summ differs from these approaches by treating the embedding space probabilistically rather than deterministically. Instead of simply selecting sentences based on proximity to a centroid or other heuristics, Vec2Summ models the full distribution of embeddings and samples from this distribution to capture both central tendencies and semantic diversity. By modeling the semantic space as a multivariate Gaussian distribution, Vec2Summ captures both the central tendency and the variance of the document collection, enabling more representative sampling that better reflects the diversity of the original content. 

\section{Vec2Summ Framework}
\subsection{Limitations of Previous Approaches and Our Solution}
Recent advances in LLM-based summarization have yielded strong results, particularly in single-document or short-input scenarios. However, these systems remain fundamentally limited by context window constraints, high inference costs, and susceptibility to information dilution in large or noisy collections. As noted by \citet{liu_lost_2024}, LLMs can easily lose coherence or relevance when processing long, multi-source inputs. This is especially problematic in domains such as social media or customer reviews, where the goal is often to extract a broad semantic overview rather than enumerate every detail. Yet, most existing approaches continue to prioritize extractive coverage over conceptual abstraction.

Vec2Summ reframes the task by operating directly in embedding space, rather than in raw text. At the heart of this method is the observation that the \emph{mean vector} of a document collection provides a compact and expressive representation of its central semantics. This enables summarization to scale beyond traditional context boundaries while preserving interpretability and diversity. While we sample from a multivariate Gaussian centered at the mean to support embedding inversion, the distributional modeling is not intended to encode the corpus itself. Rather, it serves as a tool to generate semantically similar but distinct reconstruction candidates—much like varying temperature during text generation. This approach enables controllable sampling and fluent decoding via recent advances in embedding-to-text inversion~\cite{morris_text_2023}, yielding summaries that are both efficient and expressive across large document sets.

\subsection{Framework Overview}
The Vec2Summ approach consists of four main steps:

\subsubsection{Embedding Generation}
Given a collection of input texts $\mathcal{D} = \{d_1, d_2, \ldots, d_n\}$, we first preprocess each document using standard text cleaning techniques to remove noise such as URLs, mentions, hashtags, and redundant whitespace. We then compute dense vector representations (embeddings) for each document using either OpenAI's text-embedding-ada-002 model or the General Text Representations (GTR) model~\cite{ni_large_2021}. This results in a set of embeddings $\mathcal{E} = \{e_1, e_2, \ldots, e_n\}$ where $e_i \in \mathbb{R}^d$ and $d$ is the dimensionality of the embedding space.

\subsubsection{Distribution Modeling}
We model the semantic space of the document collection as a multivariate Gaussian distribution. We compute the mean vector $\mu$ and covariance matrix $\Sigma$ of the embeddings:

\begin{align}
\mu &= \frac{1}{n}\sum_{i=1}^{n} e_i \\
\Sigma &= \frac{1}{n}\sum_{i=1}^{n} (e_i - \mu)(e_i - \mu)^T
\end{align}

To ensure numerical stability, we verify that the covariance matrix is positive definite (which should always be true) by checking its eigenvalues. If any eigenvalue is negative or close to zero (below a threshold $\epsilon = 10^{-6}$), we apply regularization:

\begin{align}
\Sigma_{reg} = \Sigma + \lambda I
\end{align}

where $\lambda$ is a small positive constant (typically $10^{-4}$) and $I$ is the identity matrix. This regularization ensures that the covariance matrix remains well-conditioned for sampling. Our experiments showed that a single Gaussian offers a good balance between model complexity and performance for most use cases.

\subsubsection{Sampling from Distribution}
We sample $k$ vectors from the multivariate Gaussian distribution $\mathcal{N}(\mu, \Sigma)$:

\begin{align}
\hat{e}_j \sim \mathcal{N}(\mu, \Sigma) \quad \text{for } j = 1, 2, \ldots, k
\end{align}

These sampled vectors $\hat{\mathcal{E}} = \{\hat{e}_1, \hat{e}_2, \ldots, \hat{e}_k\}$ represent points in the semantic space that capture both the central tendency (via the mean) and the variance (via the covariance matrix) of the original document collection.

We also explored temperature-controlled sampling by scaling the covariance matrix:

\begin{align}
\hat{e}_j \sim \mathcal{N}(\mu, T \cdot \Sigma)
\end{align}

where $T$ is a temperature parameter. Higher values of $T$ lead to more diverse samples, while lower values concentrate samples closer to the mean. In our experiments, we found that $T = 1.2$ provides a good balance between diversity and relevance.

\subsubsection{Text Reconstruction}
We reconstruct text from each sampled embedding using the vec2text framework~\cite{morris_text_2023}, which inverts embeddings back into natural language. The vec2text framework consists of two main components: a hypothesizer and a corrector.

The hypothesizer $H$ generates an initial text candidate $\hat{d}_j^{(0)}$ given a sampled embedding $\hat{e}_j$:

\begin{align}
\hat{d}_j^{(0)} = H(\hat{e}_j)
\end{align}

The corrector $C$ then iteratively refines this candidate to better match the target embedding:

\begin{align}
\hat{d}_j^{(t+1)} = C(\hat{d}_j^{(t)}, \hat{e}_j)
\end{align}

for $t = 0, 1, \ldots, T-1$, where $T$ is the number of correction iterations (typically 3-5).

The corrector operates by computing the embedding of the current candidate, comparing it to the target embedding, and making targeted edits to reduce the distance between them. The objective function for the corrector is:

\begin{align}
\mathcal{L}(\hat{d}, \hat{e}) = ||E(\hat{d}) - \hat{e}||_2^2 + \lambda \cdot \text{PPL}(\hat{d})
\end{align}

where $E$ is the embedding function, $\text{PPL}$ is a language model perplexity term that ensures fluency, and $\lambda$ is a hyperparameter that balances embedding fidelity and text fluency.

After $T$ iterations, we obtain the final reconstructed text $\hat{d}_j = \hat{d}_j^{(T)}$. This process is repeated for each sampled embedding, resulting in a set of reconstructed texts $\hat{\mathcal{D}} = \{\hat{d}_1, \hat{d}_2, \ldots, \hat{d}_k\}$.

\subsection{Algorithm}
Algorithm~\ref{alg:Vec2Summ} provides the pseudocode for the Vec2Summ framework.

\begin{algorithm}
  \caption{Vec2Summ: Text Summarization via Probabilistic Sentence Embeddings}
  \label{alg:Vec2Summ}
  \begin{algorithmic}[1]
    \Require Document collection $\mathcal{D}=\{d_1,\dots,d_n\}$, \\
             number of samples $k$, embedding model $E$, \\
             vec2text model $(H,C)$, summary model $S$, \\
             temperature $T$
    \Ensure Summary of document collection
    \State // Embedding Generation
    \For{$i\gets1$ to $n$}
      \State $e_i\gets E(d_i)$
    \EndFor
    \State $\mathcal{E}\gets\{e_1,\dots,e_n\}$
    \State // Distribution Modeling
    \State $\mu\gets\frac1n\sum_{i=1}^n e_i$
    \State $\Sigma\gets\frac1n\sum_{i=1}^n(e_i-\mu)(e_i-\mu)^T$
    \State $\lambda_{\min}\gets$ smallest eigenvalue of $\Sigma$
    \If{$\lambda_{\min}<\epsilon$}
      \State $\Sigma\gets\Sigma+(\epsilon-\lambda_{\min}+\delta)\,I$
    \EndIf
    \State // Sampling
    \State $\hat{\mathcal{E}}\gets\emptyset$
    \While{$|\hat{\mathcal{E}}|<k$}
      \State $\hat e\sim\mathcal N(\mu,\,T\Sigma)$
    \EndWhile
    \State // Text Reconstruction
    \ForAll{$\hat e_j\in\hat{\mathcal{E}}$}
      \State $\hat d_j^{(0)}\gets H(\hat e_j)$
      \For{$t=0$ to $T-1$}
        \State $\hat d_j^{(t+1)}\gets C(\hat d_j^{(t)},\,\hat e_j)$
      \EndFor
      \State $\hat{\mathcal{D}}\gets\hat{\mathcal{D}}\cup\{\hat d_j^{(T)}\}$
    \EndFor
    \State // Summary Generation
    \State $summary\gets S(\hat{\mathcal{D}})$
    \State \Return $summary$
  \end{algorithmic}
\end{algorithm}

The time complexity of Vec2Summ is dominated by the embedding generation step, which is $O(n)$ where $n$ is the number of documents. The distribution modeling step is $O(nd^2)$ where $d$ is the dimensionality of the embedding space. The sampling and text reconstruction steps are $O(kd^2)$ and $O(kT)$ respectively, where $k$ is the number of samples and $T$ is the number of correction iterations. Since typically $k \ll n$, the overall complexity is $O(n + nd^2)$, which scales linearly with the number of documents.

\section{Experiment setting}
\subsection{Data}
\label{sec:data}

As previously noted, our method is particularly effective for multi-document summarization tasks, especially in domains featuring short documents such as social media posts or product reviews. To assess its performance, we conducted experiments on four datasets spanning social media, e-commerce reviews, and medical description forums. For simplicity, we refer to each individual post (from Reddit or Twitter) or review as \textit{document}.

\subsubsection{Twitter 2020 Census Corpus}
The \textbf{general} corpus comprises
\textasciitilde3.5 M English tweets posted between 1~Jan–17~Sep~2020 that mention the \textit{2020~U.S.\ Census}, the \textit{U.S.\ Census Bureau}, or one of its field surveys, harvested via the Sprinklr vendor. 

To make the dataset more manageable and also more focused, four focused slices—\textit{Biden}, \textit{Trump}, \textit{Illegal}, and \textit{Citizen}—were obtained by keyword filtering to probe topic homogeneity effects.

\subsubsection{Amazon Book Reviews}
We used the Kaggle “Amazon Reviews” dump\footnote{https://www.kaggle.com/datasets/kritanjalijain/amazon-reviews} and retained only rows whose top-level category is \textit{Books}.  
A further slice (\textit{Amazon-Books}) contains all reviews of the single most-reviewed book (Harry Potter and the Order of the Phoenix) in the collection.  

\subsubsection{Reddit TIFU}
The Reddit-TIFU dataset~\citep{kim_abstractive_2019} contains 122,933 informal “Today I F***ed Up” posts, each paired with an author-written TL;DR. In our experiments, we used the self-text (not TL;DR) as the source for the summarization task.

\subsubsection{Reddit AskDocs}
 \textit{AskDocs} is a consumer health Q\&A forum on Reddit. We used a publicly available GitHub dump\footnote{\url{https://github.com/fostiropoulos/AskDoc?tab=readme-ov-file}} containing question–answer pairs. Since this study focuses on user-generated content, we concentrate on the question side of the data. A symptom-specific subset, \textit{AskDocs-Fever}, includes only questions that mention the term \textit{fever}.

\subsubsection{Sampling Grid and Licensing}
For every dataset we draw stratified samples of $\{50,100,200,500,1000,5000,10000\}$ documents; any cell left blank in the following figures reflects an insufficient corpus size (e.g., \textit{AskDocs-Fever}).  
Third-party corpora are distributed under permissive research licences.\footnote{Our derived Twitter corpus will be released as tweet IDs in compliance with platform policy.}

\subsection{Implementation}
We implemented Vec2Summ in Python, leveraging several libraries for different components of the pipeline\footnote{The detailed code will be shared upon acceptance}:

\begin{itemize}
    \item \textbf{Embedding Generation:} The vec2text framework supports two embedding models: OpenAI’s text-embedding-ada-002 (1536 dimensions) and the General Text Representations (GTR) model (768 dimensions). Since the GTR model was trained on outputs with a fixed and truncated length, we used OpenAI embeddings in this study for both word embedding and text reconstruction.
    
    \item \textbf{Distribution Modeling:} We use NumPy for computing the mean vector and covariance matrix, as well as for checking the eigenvalues of the covariance matrix to ensure positive definiteness. 
    
    \item \textbf{Sampling:} We use NumPy's \texttt{random.multivariate\_normal} function to sample from the multivariate Gaussian distribution.
    
    \item \textbf{Text Reconstruction:} We use the vec2text library to invert embeddings back into natural language. For OpenAI embeddings, we use the pretrained corrector model "vec2text/ada-002-corrector" with 5 correction iterations. We set the maximum generated text length to 128 tokens and use a beam size of 4 for the hypothesizer.
    
    \item \textbf{Summary Generation:} We use OpenAI's GPT-4.1 model for summarize the reconstructed text from previous steps, and the prompt can be found in the Appendix. We use a temperature of 0.7 and a maximum token length of 1024 for the summary generation.
\end{itemize}

\section{Evaluation}

We evaluate the Vec2Summ framework along two dimensions: \textit{reconstruction quality} and \textit{summary quality}. Our goal is to assess both the semantic fidelity of the reconstructed texts and the effectiveness of the final summaries in capturing key information from the original corpus.

\subsection{Reconstruction Quality}
To evaluate the semantic preservation in reconstructed texts, we compare the reconstructed outputs to the original texts using cosine similarity in the embedding space. We use the text-embedding-ada-002 model from OpenAI. The embeddings for both the original and reconstructed texts are computed, and pairwise cosine similarities are calculated.

For each original text, we record the maximum similarity with any reconstructed text, and summarize these scores using the mean similarity of all pair-wise maximum similarities between reconstructed texts with each original text (i.e., the document).

This analysis provides insight into how closely the reconstructed sentences align semantically with the input distribution.

\subsection{Comparison with Direct LLM Summarization}
To benchmark the performance of Vec2Summ, we generated summaries directly from the original texts (documents) using GPT-4.1 (the prompt we use can be seen in the Appendix) and compared them to Vec2Summ outputs using the G-Eval framework\cite{liu_g-eval_2023}. Due to context window limitations, we experimented with sample sizes of 50, 100, 200, 500, and 1000, excluding larger sizes such as 5000 and 10,000. The upper limit was set to 1000 to ensure feasibility within the context constraints.

The comparison was based on four key criteria: (1) Coverage, (2) Conciseness, (3) Coherence, and (4) Factual Accuracy.

Each summary received an overall score on a 1–5 scale, along with detailed qualitative feedback, enabling us to evaluate the relative strengths and weaknesses of Vec2Summ in comparison to direct LLM-based summarization. The prompt we used for this task can be seen in the Appendix.

\subsection{Embedding Space Visualization}

To gain insight into the geometry of the sampling process, we visualize the embedding space using Principal Component Analysis (PCA). The original embeddings, sampled vectors, and reconstructed embeddings are projected into two dimensions. The resulting plots help validate that the sampled vectors occupy a semantically meaningful region near the original data manifold.

\section{Results}
\subsection{Compression and Computational Efficiency}
\begin{figure}[t]
    \centering
    \includegraphics[width=0.5\textwidth]{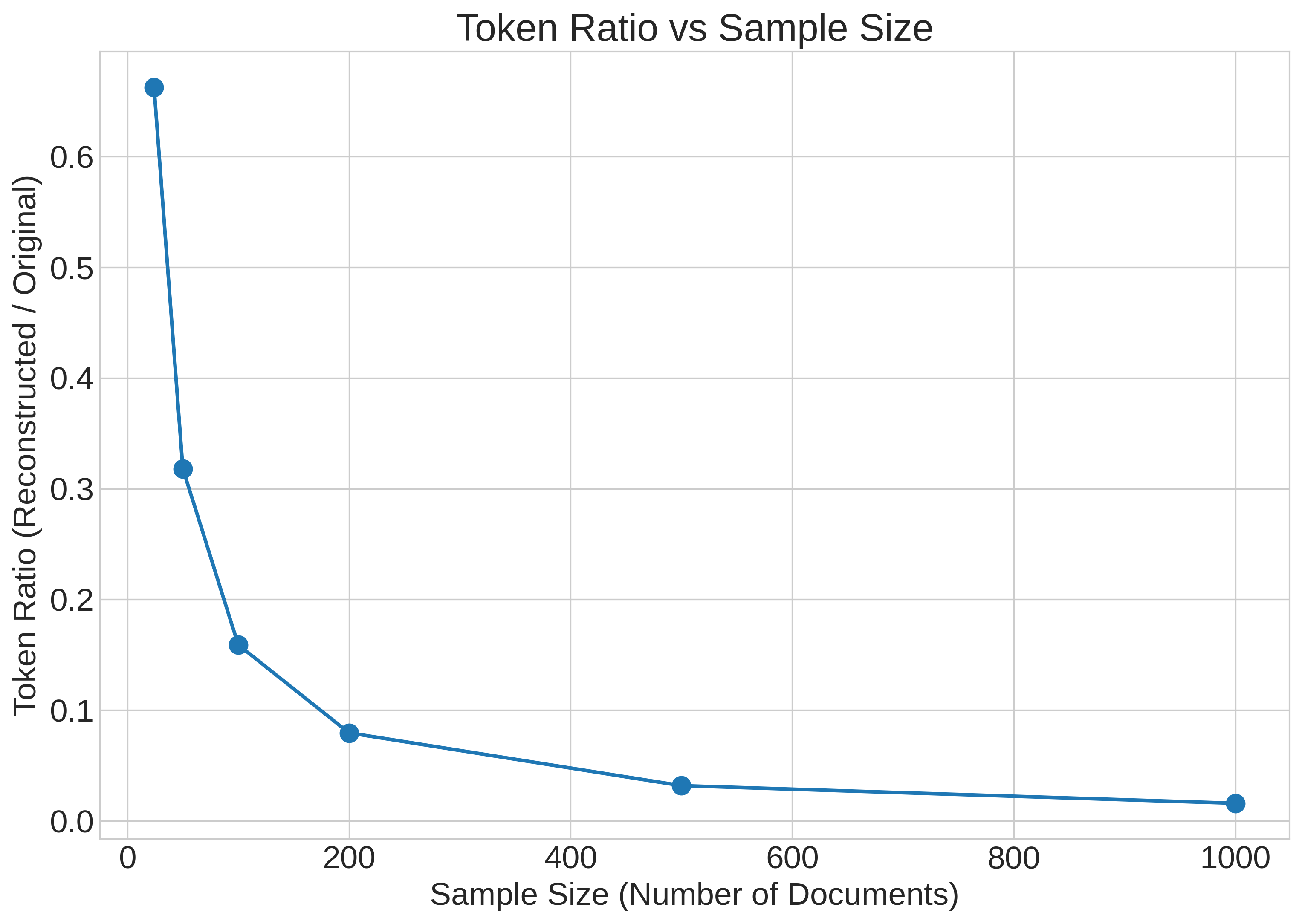}
    \caption{Computational efficiency comparison between Vec2Summ and direct LLM summarization across different corpus sizes. }
    \label{fig:efficiency_comparison}
\end{figure}
Before examining the quality of the generated summaries, we first assess the \textbf{efficiency} of Vec2Summ in comparison to direct LLM-based summarization. As discussed earlier, a core advantage of Vec2Summ is that it avoids the need to pass the entire corpus through the LLM to produce a meaningful summary. Instead, it operates on a compact subset of semantically representative sentences. This efficiency gain becomes increasingly significant as the size of the input corpus grows, making the method particularly well-suited for large-scale summarization tasks.

To illustrate this, we used the \textit{citizen} dataset to estimate the compression ratio achieved by our method relative to the total number of tokens in the original corpus, as shown in Figure~\ref{fig:efficiency_comparison}. Vec2Summ exhibits strong compression capabilities that scale effectively with corpus size. For instance, when summarizing a corpus of 100 documents, Vec2Summ reduces the content to just 15\% of the original token count. As the corpus size grows to 1,000 documents, the compression ratio improves dramatically—up to 99.9\%—while the parameter count remains fixed at $O(d + d^2)$.

\subsection{Semantic Representation Quality}
\begin{figure}
    \centering
    \includegraphics[width=\linewidth]{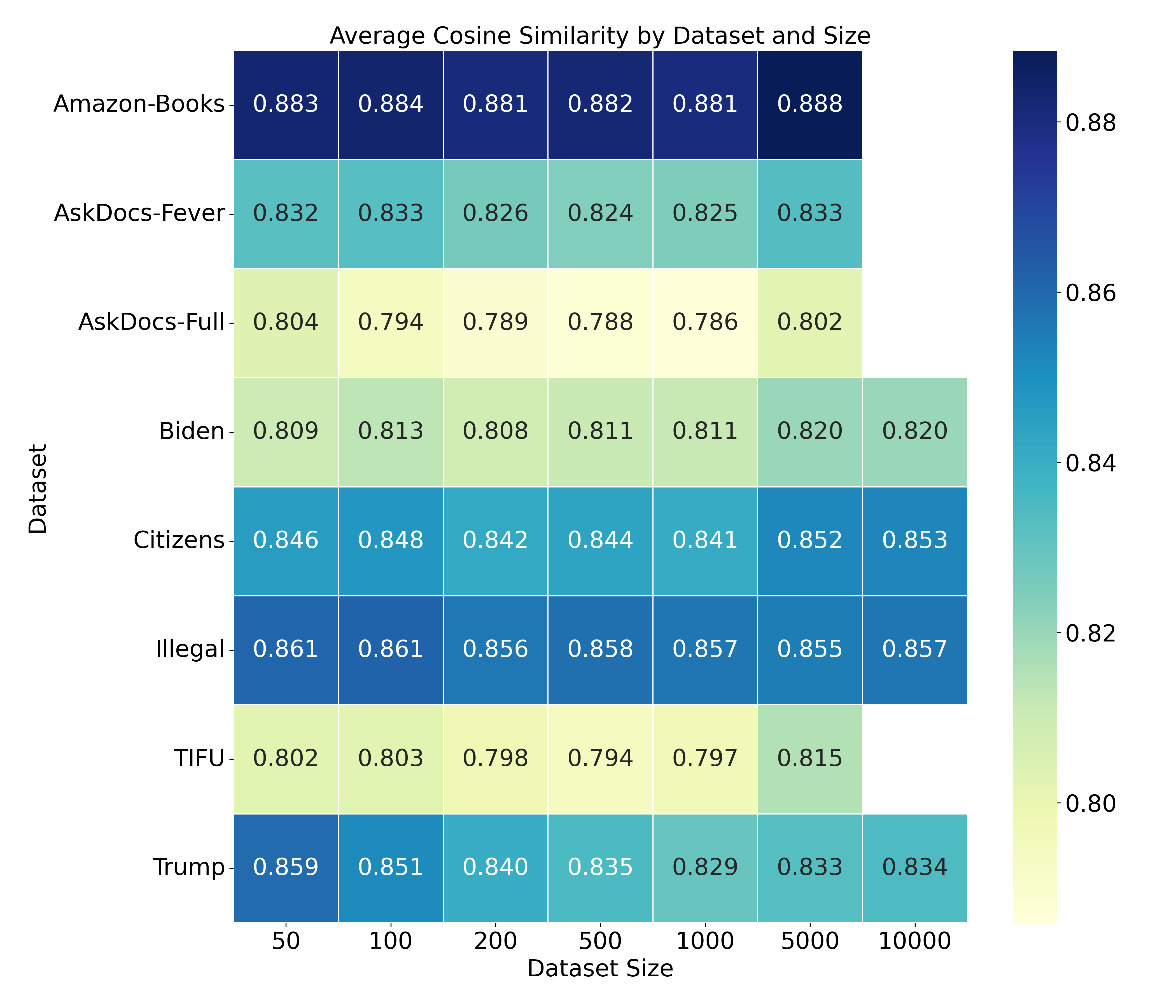}
    \caption{Average pair-wise mean cosine similarity between each reconstructed text and its corresponding source text (lighter = lower, darker = higher). }
    \label{fig:cos}
\end{figure}

Figure~\ref{fig:cos} presents a heat-map of the average pair-wise mean cosine similarity between each reconstructed text and its corresponding source text, stratified by dataset (rows) and by the number of source passages used to build the reconstruction model (columns: 50 → 10 000). Some datasets lack a 10,000 sample entry because the original corpora simply do not contain that many distinct documents. Across all 56 (8 datasets × 7 sizes) experimental conditions, the similarity remains well above 0.78; 46/56 cells exceed 0.82, corroborating the strong semantic faithfulness already noted above.

The highest fidelity is observed on  \textit{Amazon-Books} (0.883–-0.888) and  \textit{Illegal} (0.855--0.861), suggesting that our method performs particularly well with narrative reviews and statutory language. The lowest—though still notable—scores appear on \textit{AskDocs-Full} (minimum 0.786), a forum characterized by medical jargon and long-tail personal stories. This indicates that our method is less suited for cases where each document represents a unique instance, with limited cross-document correlation within the corpus. The remaining datasets—\textit{Citizen}, \textit{Trump}, \textit{AskDocs-Fever}, \textit{Biden}, and \textit{TIFU}—form a tight cluster between 0.80 and 0.86, demonstrating robustness across political, conversational, and short-form narrative domains.

Similarity varies by at most ±0.02 within each dataset, and no monotonic degradation is visible as we scale from 50 to 10,000 training items. In fact, several datasets (\textit{Citizen}, \textit{ \textit{Amazon-Books}}) exhibit a slight uptick ($\approx +0.01 - 0.02$) at the 10 000-item mark, while others (e.g.,  \textit{Trump},  \textit{AskDocs-Fever}) recover the small dip seen at medium sizes (200--1,000). This stability indicates our reconstruction module does not trade fidelity for scale; semantic alignment saturates quickly and stays flat thereafter.

\subsection{Comparison with Direct LLM Summarization}
\begin{figure}[t]
  \centering
  \includegraphics[width=\linewidth]{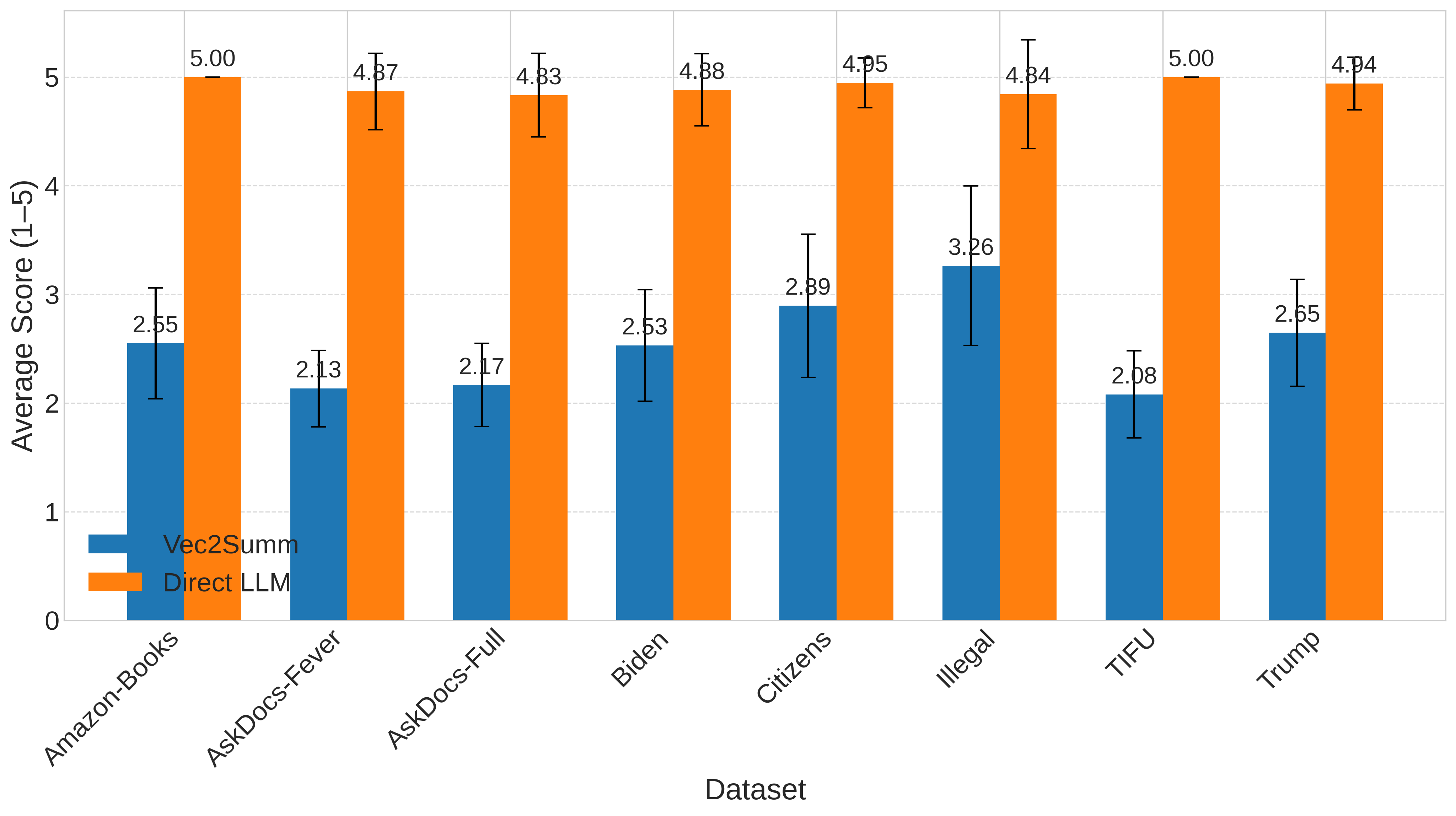}
  \caption{Average G-Eval scores (1–-5 scale) for Vec2Summ and direct GPT-4
           summarization across eight datasets. Error bars indicate 95\%
           confidence intervals.}
  \label{fig:performance_comparison}
\end{figure}

We conducted a comprehensive evaluation comparing \textbf{Vec2Summ} against a \textbf{direct LLM summarization} baseline to assess both output quality and scalability. As shown in Figure~\ref{fig:performance_comparison}, we report average G-Eval scores across eight datasets using a fixed sample size of 1,000 documents. Each experiment was repeated five times to ensure stability, and the results were averaged across runs.

As we mentioned, summaries were evaluated on four criteria — coverage, factual accuracy, coherence, and conciseness — with final scores reported on a 1–5 scale. To provide additional insight into the qualitative differences between the two methods, we include a representative example from the \textit{Citizen} dataset (sample size 1,000) in the Appendix, illustrating how Vec2Summ and the direct LLM approach diverge or are similar in structure and focus.

As shown, the direct LLM approach consistently outperforms Vec2Summ in
terms of raw G-Eval scores, often reaching near-perfect evaluations
(4.8--5.0). Vec2Summ, while lower, still maintains a respectable range of
scores (2.1--3.3), indicating that it is capable of generating coherent and
factually consistent summaries, even at scale.

We emphasize that these results should be interpreted with caution when applied
to large input sizes. As the number of documents to summarize increases (e.g.,
1,000 documents), the task becomes more cognitively demanding for any
evaluator—including both humans and G-Eval—to consistently judge the quality of
a summary. Thus, while the scoring provides a useful benchmark, it may not fully
capture consistency or depth of justification at higher volumes.

Another key observation is the difference in summary length and structure.
Vec2Summ outputs are typically concise—consisting of a single paragraph—while
direct LLM summaries often span multiple paragraphs and include more detailed
elaboration. This verbosity allows the direct LLM approach to score higher on
the \textit{coverage} criterion, which heavily influences the G-Eval composite
score. In contrast, Vec2Summ focuses on high-density content selected via
embedding similarity, producing shorter summaries that may omit low-frequency or
peripheral details. This difference is also evident in the example provided in the Appendix.

Indeed, the most common critique in G-Eval comments on Vec2Summ outputs was
“limited coverage.” However, we argue that this property is not necessarily a
flaw. In many practical applications—such as dashboards, monitoring tools, or
first-pass summarization—the main goal is to convey the \textit{gist} of a large
set of documents rather than enumerate every specific point. From this perspective,
Vec2Summ offers a valuable trade-off: reduced length and faster readability, in
exchange for a moderate loss in detail.

In summary, although direct LLM summarization remains stronger under
coverage-heavy metrics, our findings show that Vec2Summ delivers competitive
performance in a more compact format. These results are encouraging and suggest
that Vec2Summ can scale to even larger datasets while still providing useful,
interpretable summaries.

\subsection{Embedding Space Preservation}
\label{sec:embedding}

\begin{figure}[t]
  \centering
  \includegraphics[width=\linewidth]{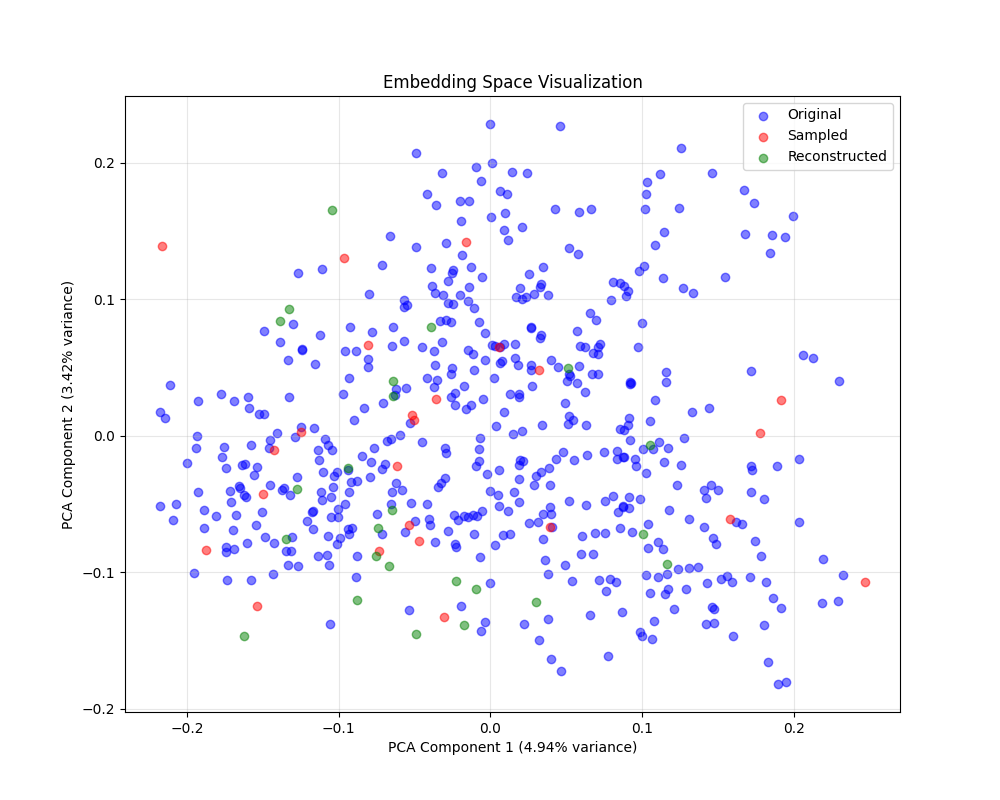}
  \caption{PCA projection of embedding space from the \textit{Biden} dataset
           using 500 sampled documents. Blue = original embeddings;
           red = sampled subset; green = reconstructed texts.
           Sampled points span the semantic space well,
           and reconstructions remain close to their targets.}
  \label{fig:embedding_distribution}
\end{figure}

To assess the semantic similarity of our sampled and reconstructed texts, we
visualized their embeddings using PCA projection onto two components. As shown
in Figure~\ref{fig:embedding_distribution}, the sampled embeddings (in red)
retain the structure of the full embedding distribution (in blue), indicating
that the selection strategy preserves the major semantic modes of the corpus.

The reconstructed text embeddings (in green) cluster tightly around their sampled
counterparts, suggesting that the text-to-text generation process in Vec2Summ
successfully captures the intent and meaning of the sampled inputs. This figure
uses the \textit{Biden} dataset with 500 sampled documents as an illustrative
example, but similar patterns were observed across other domains and sample
sizes.

One caveat is that not every reconstructed point lands precisely on its sampled
source in embedding space. While the reconstructions are close enough to retain
semantic proximity, perfect recovery may require domain-specific fine-tuning of
the embedding-to-text decoder. Nonetheless, the spatial consistency observed
here highlights that Vec2Summ can approximate large-scale semantic coverage with
significantly fewer tokens and minimal distortion.

\section{Discussion}

The results demonstrate that Vec2Summ, by leveraging mean vector encoding in semantic space, effectively captures the semantic diversity of document collections while preserving coherence in the generated summaries. Its probabilistic sampling approach allows for the inclusion of information distributed throughout the semantic space of the corpus, rather than concentrating solely on the most central or frequently occurring content.

Vec2Summ also presents a compelling alternative to direct LLM-based summarization, particularly for large, topically focused corpora where efficiency and scalability are critical. By sampling across the full semantic distribution—including its tails—Vec2Summ is able to surface diverse and less redundant information that may be overlooked by direct LLM summarization methods, which often emphasize more central content.

The performance trade-offs observed in our experiments suggest that Vec2Summ is best suited for tasks where capturing the general semantic content is more important than preserving fine-grained details. Its efficiency gains become increasingly significant as the corpus size grows, making Vec2Summ particularly useful for summarizing large collections that exceed the context window of typical LLMs.

That said, we acknowledge several limitations and areas for future improvement. First, while the mean vector representation of a semantic space works well for multi-document summarization, it may not generalize to other domains. For instance, we conducted pilot studies on book and news article summarization, but neither yielded results meaningful enough to include in this paper.

Additionally, our observations suggest that Vec2Summ performs best when applied to datasets with two key characteristics: (1) order invariance—where shuffling the document order does not affect the overall content—and (2) topical coherence—where the documents share a common theme or event. Under these conditions, the mean vector and multivariate distribution modeling help mitigate the isotropy problem commonly observed in embedding spaces.

Finally, even with an off-the-shelf setup, we are able to generate coherent and informative summaries from the mean vector. However, we believe performance can be further improved by fine-tuning the vec2text decoder on domain-specific text. While this would require additional effort to train, the process can be carried out in a self-supervised manner, meaning it does not require labeled external data—only raw text from the target domain.

\section{Appendices}
\subsection{Prompt for generating Vec2Summ summarization}
    You are given a set of text fragments that represent different aspects of a document collection.
    Your task is to generate a comprehensive summary that captures the key information present in these fragments.
    Focus on identifying common themes, important details, and diverse perspectives.
    
    Text fragments:
    \{reconstructed\_texts\}
    
    Summary:
\subsection{Prompt for direct LLMs summarization}
        Please summarize the following collection of texts. Focus on capturing the main points, 
        key information, and overall sentiment. The summary should be comprehensive yet concise.
        
        Texts to summarize:
        \{source\_text\}  
\subsection{G-Eval prompt}
        You are an expert evaluator for text summarization systems.
        
        I'll provide you with a source document and a summary. Your task is to evaluate how well the summary covers the important information in the source document.
        
        Source document:
        \{source\_text\}
        
        Summary:
        \{summary\}
        
        Please evaluate the COVERAGE of this summary on a scale of 1-5, where:
        1: The summary misses almost all important information
        2: The summary misses most important information
        3: The summary covers about half of the important information
        4: The summary covers most important information
        5: The summary covers all or almost all important information
        
        First, identify the key points in the source document.
        Then, check which of these key points are covered in the summary.
        Finally, provide your rating and explanation.
        
        Your response should follow this format:
        Key points in source: [list key points]
        Points covered in summary: [list covered points]
        Points missing in summary: [list missing points]
        Coverage score: [1-5]
        Explanation: [your explanation]

\subsection{Direct LLM summary vs Vec2Summ summary}
\subsubsection{Direct LLM summary}
The collection of texts centers on widespread debate and anxiety among politically engaged Americans regarding the 2020 US presidential election, with a particular emphasis on the unique significance of it being a census year. The central themes include concerns about Joe Biden’s candidacy, the Democratic Party’s prospects, and the long-term implications of the election for political power and representation.

A major focus is the importance of the 2020 census, which will determine congressional redistricting and potentially shape the balance of power for the next decade. Many contributors stress that control of state legislatures and Congress is crucial, as these bodies will oversee the redrawing of district maps. There is significant apprehension that a Republican victory in a census year could entrench GOP dominance through gerrymandering, affecting policy, representation, and the judiciary for years to come.

Joe Biden is frequently discussed as the likely Democratic nominee. While some view him as a pragmatic choice to prevent further Republican control and protect key issues such as healthcare, civil rights, and the Supreme Court, others express skepticism about his ability to energize young, progressive, and Latino voters. Concerns are raised about his record, cognitive abilities, and potential to depress turnout, drawing comparisons to the 2016 election. There is also criticism of establishment politics and frustration with the Democratic Party’s strategy.

The texts highlight the importance of down-ballot races, emphasizing that winning the presidency alone is insufficient; control of the Senate, House, and state legislatures is also necessary to influence redistricting and policy outcomes. Voter turnout, demographic shifts, and the mobilization of key groups—particularly young voters, African Americans, and Latinos—are seen as critical factors in determining the election’s outcome.

Additional concerns include allegations of corruption (notably involving Hunter Biden), voter suppression, misinformation, and manipulation of both the census and the electoral process. There is mutual distrust between parties, with accusations of unethical tactics and fears of primary rigging.

Overall, the sentiment across the texts is one of urgency, frustration, and strategic calculation. There is deep skepticism about both major candidates, but a clear recognition that the structural stakes of the 2020 election—especially due to the census and its impact on redistricting and the courts—make its outcome particularly consequential for the future of American democracy and political representation.
\subsubsection{Vec2Summ summary}
The collection of texts presents a fragmented and chaotic discussion about American politics, focusing on elections, the census, and party dynamics. The main themes include confusion and frustration over the electoral process, skepticism about the census and its impact, and criticism of both Democratic and Republican figures. There are repeated references to prominent politicians like Bernie Sanders, Joe Biden, John McCain, Barack Obama, Donald Trump, and others, often in contradictory or nonsensical contexts.

The sentiment is generally negative, with expressions of distrust toward the political system, the census, and government institutions like the EPA and the Federal Reserve. There is also a sense of cynicism about political campaigns, polling, and the effectiveness of both parties. Some texts mention issues of race and representation, as well as concerns about economic conditions and joblessness.

Overall, the texts reflect disillusionment, confusion, and a lack of faith in political processes, with scattered calls for action but little coherent direction. The tone is often sarcastic or mocking, and the content jumps between topics without clear resolution.
\label{sec:bibtex}

\bibliography{acl_latex}

\end{document}